\documentclass[12pt,a4paper]{article}
\usepackage[utf8]{inputenc}
\usepackage{graphicx}
\usepackage{amsmath}
\usepackage{amssymb}
\usepackage{authblk}
\usepackage{etoolbox}
\usepackage{hyperref}
\usepackage{xcolor}
\usepackage{tabularx}
\usepackage{booktabs}
\usepackage{array}
\usepackage{multirow}
\usepackage{makecell}
\usepackage{indentfirst}
\usepackage[T1]{fontenc}
\usepackage{newtxtext} 
\usepackage{newtxmath}
\usepackage{geometry}
\usepackage{caption}
\usepackage{float}

\date{}

\begin{document}
\title{Automated Review Generation Method Based on Large Language Models}

\author[1\#]{Shican Wu}
\author[1\#]{Xiao Ma}
\author[1]{Dehui Luo}
\author[1]{Lulu Li}
\author[2]{Xiangcheng Shi}
\author[2]{Xin Chang}
\author[1]{Xiaoyun Lin}
\author[2]{Ran Luo}
\author[1,3]{Chunlei Pei}
\author[4]{Changying Du}
\author[1,5,*]{Zhi-Jian Zhao}
\author[1,2,3,5,6,7,8,*]{Jinlong Gong}

\affil[1]{School of Chemical Engineering and Technology; Key Laboratory for Green Chemical Technology of Ministry of Education, Tianjin University; Collaborative Innovation Center of Chemical Science and Engineering (Tianjin); Tianjin 300072, China}
\affil[2]{Joint School of National University of Singapore and Tianjin University, International Campus of Tianjin University, Binhai New City, Fuzhou 350207, Fujian, China}
\affil[3]{Zhejiang Institute of Tianjin University Ningbo, Zhejiang 315201, China}
\affil[4]{AIStrucX Technologies, No. 26, Information Road, Haidian District, Beijing 100000, China}
\affil[5]{International Joint Laboratory of Low-carbon Chemical Engineering, Tianjin 300192, China}
\affil[6]{Haihe Laboratory of Sustainable Chemical Transformations, Tianjin 300192, China}
\affil[7]{National Industry-Education Platform of Energy Storage, Tianjin University, 135 Yaguan Road, Tianjin 300350, China}
\affil[8]{Tianjin Normal University, Tianjin 300387, China}
\affil[*]{Corresponding authors. Email: zjzhao@tju.edu.cn; jlgong@tju.edu.cn}
\affil[ \#]{These authors contributed equally: Shican Wu, Xiao Ma}
\renewcommand\Authands{ and }
\maketitle
	
	\begin{abstract}
		Literature research, vital for scientific work, faces the challenge of surging information volumes exceeding researchers' processing capabilities. We present an automated review generation method based on large language models (LLMs) to overcome efficiency bottlenecks and reduce cognitive load. Our statistically validated evaluation framework demonstrates that the generated reviews match or exceed manual quality, offering broad applicability across research fields without requiring users' domain knowledge. Applied to propane dehydrogenation (PDH) catalysts, our method swiftly analyzed 343 articles, averaging seconds per article per LLM account, producing comprehensive reviews spanning 35 topics, with extended analysis of 1041 articles providing insights into catalysts' properties. Through multi-layered quality control, we effectively mitigated LLMs' hallucinations, with expert verification confirming accuracy and citation integrity while demonstrating hallucination risks reduced to below 0.5\% with 95\% confidence. Released Windows application enables one-click review generation, enhancing research productivity and literature recommendation efficiency while setting the stage for broader scientific explorations.
	\end{abstract}
	
	
	\pagebreak
	\section{INTRODUCTION}
	
	Peer-reviewed academic literature functions as a critical medium for scientific knowledge dissemination, enabling researchers to advance human understanding through cumulative progress\cite{1}. The clarity and rigor of scientific language facilitates entity description, concept extraction, and consensus building, ensuring cognitive consistency between information senders and receivers. However, the exponential growth in publications has exceeded researcher' processing capacity\cite{2,3,4}, necessitating efficient tools for literature analysis and integration, thus avoiding redundant discoveries and broadening research perspectives.
	
	Natural language processing (NLP), encompassing co-reference resolution, semantic analysis, etc.\cite{5}, powers literature comprehension. Since November 2022, Large Language Models like ChatGPT, as the latest NLP advancement, have exhibited unprecedented language understanding\cite{6}. Leading LLMs have surpassed human performance on various benchmarks including MMLU\cite{7}, which tests undergraduate knowledge, and GPQA Diamond\cite{8}, which assesses graduate-level reasoning, positioning them as potential "second brains" for efficient literature processing\cite{6,9}. Recent studies like PaperQA\cite{10} and its improved version PaperQA2\cite{11}, as retrieval-augmented generation (RAG\cite{12}) agents, demonstrate excellent performance in literature-related tasks including retrieval, question-answering, summarization, and contradiction detection, surpassing human expertise in some aspects. AcademicGPT\cite{13} provides comprehensive research support, while CuriousLLM\cite{14} enhances multi-document question-answering through reasoning-based traversal agents. However, these applications require user-provided literature, rely on question-answer interactions, or focus on specific points, limiting their transferability.
	
	The review format effectively integrates literature information and generalizes across disciplinary fields, naturally leading to automated review generation research. However, early attempts encountered several limitations. They either reduced reviews to multi-document summarization\cite{15} or depended on existing reviews and citation networks\cite{16,17,18,19,20}, which struggled with rapidly evolving fields and underestimated recent publications due to citation lag. Additional constraints included using only abstracts instead of full texts as input data\cite{17,18,19,20} and employing either extractive summarization rather than integrated generation\cite{16,17,18} or template-based generation\cite{19}, risking information loss and redundancy. Recent LLM-based solutions include: multi-AI agent systems\cite{21} for full-process automation from research question generation to data extraction; LitLLM\cite{22} combining RAG with LLM reranking to generate high-quality literature reviews based on user-provided abstracts; LLAssist\cite{23} and related work\cite{24} for literature screening; and ChatCite\cite{25} improving summary quality through human-like workflows. These advances enhance automated review generation while enabling efficient academic research.
	
	Based on the potential of LLMs, this study proposes an automated review generation method based on LLMs, builds an end-to-end data pipeline from literature retrieval to final review text generation. By leveraging information refinement and knowledge construction capabilities of LLMs, this method overcomes human cognitive limitations in single-threaded processing and memory capacity, reducing researchers' cognitive load while offering superior speed and scalability, thereby substantially conserving professional human resources. However, two critical challenges persist: the macro-level requirement for systematic quality evaluation and comparison with manual reviews, and the micro-level necessity to effectively mitigate LLM hallucinations. To address these challenges, we designed a dual-baseline automatic evaluation framework with rigorous statistical validation, alongside multi-level quality control strategies throughout the process. The distinctive feature of method lies in its adaptability to diverse disciplinary terminologies and knowledge structures without domain-specific training, facilitating both comprehensive field overviews for experienced researchers and accessible entry points for those lacking relevant background, opening up new possibilities for promoting interdisciplinary research and knowledge dissemination. This approach holds substantial scientific significance by enhancing literature processing efficiency, fostering knowledge discovery, and stimulating innovation. Its ability to promote interdisciplinary communication and knowledge integration positions it as a potential cornerstone of modern research infrastructure, accelerating scientific discovery and technological advancement across domains.
	
	\section{RESULTS AND DISCUSSION}
	
	\subsection{Automated retrieval}
	
	Automated review generation fundamentally relies on retrieving and extracting scientific literature, with output quality dependent on input timeliness, quality, and breadth. To demonstrate cross-disciplinary generalization without human intervention, we conducted a case study on propane dehydrogenation (PDH) catalysts, searching chemistry and chemical engineering journals (1980-2024) ranked Q1 in the Chinese Academy of Sciences journal classification on Google Scholar through SerpAPI.
	
	The automated retrieval yielded 1420 initial results from Google Scholar. To address the challenge of irrelevant or duplicate findings, we implemented a dual-level filtering process. The first level employed quick filtering of abstracts and titles to remove obviously irrelevant documents, as detailed in Method section, serving as a rapid but less precise narrowing method. The second level involved deeper LLM-based analysis of full texts, offering higher accuracy albeit at a slower pace. This coarse-to-fine screening method, reminiscent of high-throughput screening, enabled us to identify literature pertinent efficiently and accurately to our research. The initial screening shortlisted 343 articles as related to our topic. Subsequent LLM evaluation further confirmed 238 of these articles as relevant.
	
	\subsection{Implementation and analysis of one-click automated review generation}
	
	Using PDH catalysts as an example and building on the aforementioned automated retrieval, we have effectively produced high-quality, specialized review articles. Considering that the entire process is completely end-to-end without the need for human intervention, we believe that a single domain example is sufficient to demonstrate the applicability of this method. The main reason for limiting the journal range to Q1 journals is that although the impact factor of journals may not be closely related to the quality of articles, the lower limit of literature in Q1 journals that have passed strict peer review is relatively higher. Considering users lacking prior knowledge in the target domain, directly traversing Q1 chemistry journals provides an efficient starting point. We solemnly declare here that we are not encouraging users to only consider Q1 journals, but rather suggesting that in the initial stage, one can consider starting with Q1 journals, and for research on the entire field, all possibilities should be explored, which is also supported by our method. In the Windows GUI we provide, using Q1/Q2-3 journals is an optional button, allowing users to choose for themselves. For those with domain familiarity, the program allows the specification of a custom journal list to refine article selection.
	
	We evaluated two topic construction strategies: based on existing reviews (9 topics, 35 questions, 125 citations) and direct LLM generation (12 topics, 12 questions, 43 citations). The examples showcased in subsequent sections and SI are based on outlines derived from existing reviews. The content has been manually checked by experts in the relevant field, with no errors in knowledge, correct referencing of cited literature, and a length and citation count that align with conventional review standards (see SI). The method's effectiveness stems from LLMs' human-level or superior language comprehension abilities, coupled with the injection of domain knowledge from retrieved literature through context window, thereby enabling generalization across all research fields. Beyond content accuracy, the method enables customizable research focus through supporting of adding specific questions and provides forward-looking insights with comprehensive understanding sections. To facilitate broad adoption, we developed an open-source Python3 GUI enabling one-click review generation without programming expertise or domain knowledge.
	
	\subsection{Evaluation of generated review quality}
	
	Research demonstrates LLMs excel in evaluation tasks, with GPT-4 surpassing both crowdsourced workers\cite{26} and experts\cite{27} in text annotation accuracy and reliability, and bias control for complex tasks requiring contextual knowledge reasoning\cite{27}. LLMs show comparable or superior performance to human annotation in persuasiveness, accuracy, and satisfaction\cite{28}. Studies confirm the capability of LLMs to evaluate other LLMs, with verification abilities improving faster than generation quality\cite{29}. GPT-4's evaluations exceeds 80\% consistency with human reviewers\cite{30} and exceeds 85\% alignment in pairwise comparisons\cite{31}, reaching nearly 100\% agreement when performance differences are significant\cite{31}. Self-verification benefits LLM performance\cite{32}, though biases issues\cite{33}, such as position bias\cite{34}, length bias\cite{30}, and self-bias\cite{35}, persist but can be reduced through proper design\cite{36}.
	
	Given characteristic writing patterns of LLMs and potential human evaluation bias, this study employs LLM-based evaluation exclusively. Existing LLM-based LLM evaluation methods encompass scoring, comparison, selection judgment, and comprehensive description. This study introduces a dual-baseline review quality evaluation framework, to minimize potential LLM evaluation bias and quantitatively compare LLM-generated reviews with peer-reviewed expert-written content, validating reliability through statistical analysis. In our method, we segmented 14 published Q1 reviews into 89 fragments based on semantic content. Using extracted topics from these fragments and the literature cited in the original text, we generated comparative reviews using Qwen2-7b-Instruct, Qwen2-72b-Instruct, and Claude3.5Sonnet. This methodology enabled direct comparison between human experts and LLMs in writing reviews with identical literature background, establishing a rigorous benchmark for LLMs given their limitations in accessing both human accumulated domain expertise and pre-1970 undigitized literature.
	
	In the evaluation process, we compared the performance of two models of different scales from the open-source Qwen2 series (Qwen2-7b-Instruct and Qwen2-72b-Instruct) and the closed-source model Claude3.5Sonnet (Fig. 1). Evaluation employed both self-scoring and uniform scoring by Qwen2-72b-Instruct. Intraclass Correlation Coefficient (ICC) tests and Transitive Consistency Ratio (TCR) analyses confirmed high reliability for Claude3.5Sonnet (Fig. 1(a,b,e,f)) and Qwen2-72b-Instruct (Fig. 1(c,d,e,f)), meeting human evaluation standards, while Qwen2-7b-Instruct’s results necessitated substitution with Qwen2-72b-Instruct’s scoring due to insufficient reliability. This phenomenon may stem from differences in capabilities between LLMs of different scales, specifically manifested in three dimensions: world knowledge, language understanding, and logical reasoning. We believe that small models differ significantly from large models in logical reasoning ability, while world knowledge has been supplemented through context-provided literature, and language ability differences are relatively small. For details on Qwen2-7b-Instruct’s evaluation results, see SI.
	
	\begin{figure*}[htbp]
		\centering
		\includegraphics[width=0.42\textwidth]{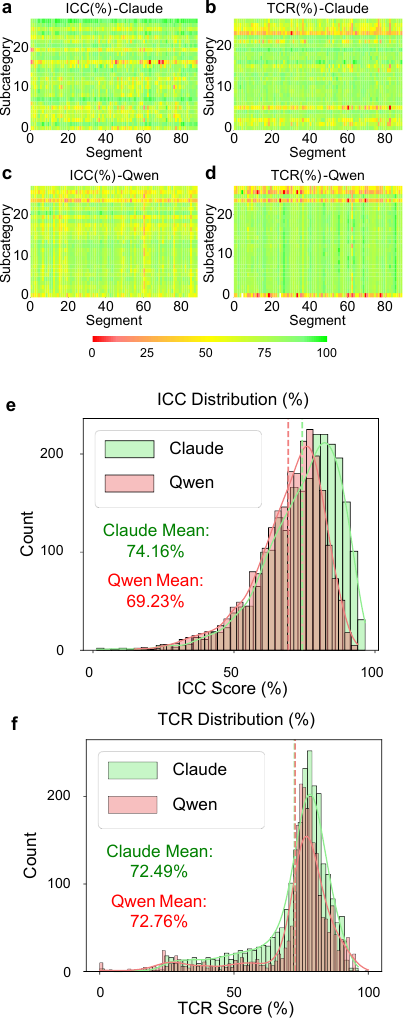}
		\caption{\textbf{Reliability verification results of the dual-baseline review quality assess- }}
		\label{fig:1}
	\end{figure*}
	
	\begin{figure*}[!h]
		\ContinuedFloat
		\caption*{
			\textbf{ment framework.} Scheme evaluation result distribution heat map, red to green showing 0\% to +100\% range, higher values indicate better performance: \textbf{a,} Intraclass correlation coefficient test for Claude3.5 Sonnet model results; \textbf{b,} Transitive consistency ratio analyses for Claude3.5 Sonnet model results; \textbf{c,} Intraclass correlation coefficient test for Qwen2-72b-Instruct model results; \textbf{d,} Transitive consistency ratio analyses for Qwen2-72b-Instruct model results. Scheme evaluation result distribution histograms: \textbf{e,} Intraclass correlation coefficient test results for Claude3.5 Sonnet model (green) and Qwen2-72b-Instruct model (red); \textbf{f,} Transitive consistency ratio analyses results for Claude3.5 Sonnet model (green) and Qwen2-72b-Instruct model (red). More data charts are available in the SI.}
	\end{figure*}
	
	Model capability significantly impacts generation quality, while our method ensures a basic lower limit of generation quality. In repeated generation tests (9 times per model), average scores, taken as comprehensive performance of models, showed that Qwen2-7b-Instruct reached 43.94\% of manual scoring, Qwen2-72b-Instruct reached 64.81\% (Fig. 2(d,f)), and Claude3.5Sonnet exceeding manual scores by 23.63\% (Fig. 2(c,f)), all significantly higher than the baseline level of direct generation. Optimal performance analysis, which taking the paragraph with the highest total score among those generated by each model as the optimal paragraph, revealed best-paragraph scores of 89.07\% for Qwen2-7b-Instruct, 92.64\% for Qwen2-72b-Instruct (Fig. 2(b,f)), and 130.79\% for Claude3.5Sonnet relative to manual scores (Fig. 2(a,f)). These results indicate multiple generations and selecting the best paragraph can improve performance of smaller models, while larger models maintain consistent quality. For optimal results, we recommend using larger models when possible; otherwise, multiple generations can enhance final performance in hardware-constrained scenarios. For details on Qwen2-7b-Instruct’s evaluation results, see SI.
		
	\clearpage
	\begin{figure*}[h!]
		\centering
		\includegraphics[width=0.5\textwidth]{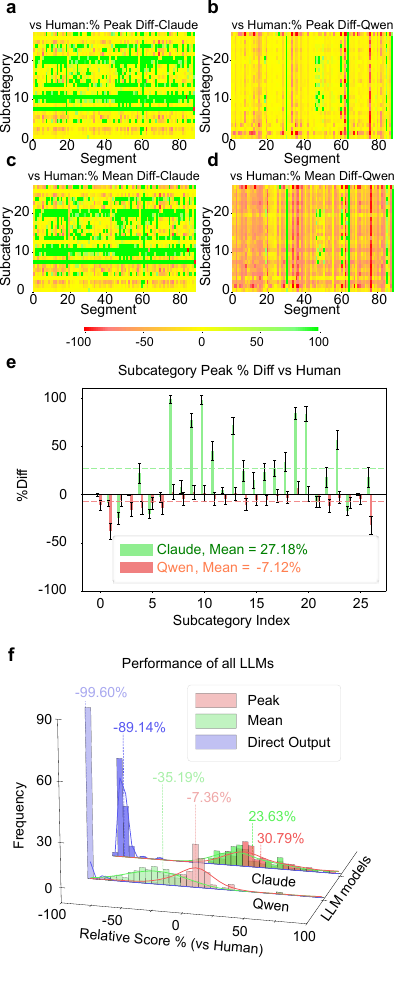}
		\caption{\textbf{Quality assessment results of automatically generated reviews.} Heat map of the percentage difference in scores of review paragraphs generated by this method relative to human scores, red to green showing -100\% to +100\% range, higher values indicate better performance, values truncated to ±100\% range, values exceeding are recorded as -100\% and +100\%: \textbf{a,} Highest scoring paragraph of Claude3.5 Sonnet model; \textbf{b,} Highest scoring paragraph of Qwen2-72b-Instruct model; \textbf{c,} Average paragraph score of Claude3.5 Sonnet model; \textbf{d,} Average paragraph score of Qwen2-72b-Instruct model. \textbf{e,} Histogram of percentage differences in scores relative to human scores for highest scoring paragraphs, average paragraph scores, and directly generated paragraph scores without going through this method for Claude3.5 Sonnet model and Qwen2-72b-Instruct model, colors ranging from dark to light representing Claude3.5 }
		\label{fig:2}
	\end{figure*}
	
		\begin{figure*}[h!]
		\ContinuedFloat
		\caption*{
			   Sonnet model and Qwen2-72b-Instruct model and Qwen2-72b-Instruct model, green, red, and blue representing highest scoring paragraphs, average paragraph scores, and directly generated result scores respectively. \textbf{f,} Bar chart of scores for highest scoring paragraphs of Claude3.5 Sonnet model and Qwen2-72b-Instruct model in each scoring sub-item, green and red representing Claude3.5 Sonnet model and Qwen2-72b-Instruct model respectively. More data charts are available in the SI.}
	\end{figure*}
		
	The near-human scores of optimal paragraphs might reflect quality-related bias[37] because their closely approximate human-level quality could make LLMs’ potential evaluation bias more prominent, while the distinctly inferior quality of direct generation remains easily distinguishable. Analysis of optimal paragraphs reveals high correlations between Qwen2-7B and Qwen2-72B across evaluation sub-items (0.926 for highest scores, 0.939 for scores relative to human benchmarks), reflecting the scaling law of LLM and suggesting potential for further improvements with model advancement (see SI for details). Our research validates the effectiveness of LLM-based automated review generation, with quality approaching or exceeding manual reviews. The effectiveness of method stems from general language processing and context learning capabilities of LLMs, rather than requiring specific domain expertise, suggests broad disciplinary applicability. While inheriting common biases and requiring readers’ professional judgment, this method serves as a supportive rather than a replacement tool for human innovation, with open-source models showing comparable capabilities to closed-source alternatives. The approach demonstrates broad cross-disciplinary potential, promising to become an important tool for promoting academic innovation and knowledge dissemination, while the dual-baseline framework offers potential methodology for evaluating LLM agent workflows where manual data acquisition is costly.
	
	\subsection{Data mining and visual analysis}
	
	Catalysts play a vital role in chemical process optimization\cite{38}, with data mining enabling accelerated design through pattern recognition\cite{39}. Analyzing 839 PDH catalyst papers from a total of 1041 articles filtered by abstracts and titles in Q1-Q3 chemistry journals (1980-2024), our data mining module revealed comprehensive insights into catalysts' composition, structure, and performance.
	
\begin{figure}[H]
		\centering
		\includegraphics[width=1\textwidth]{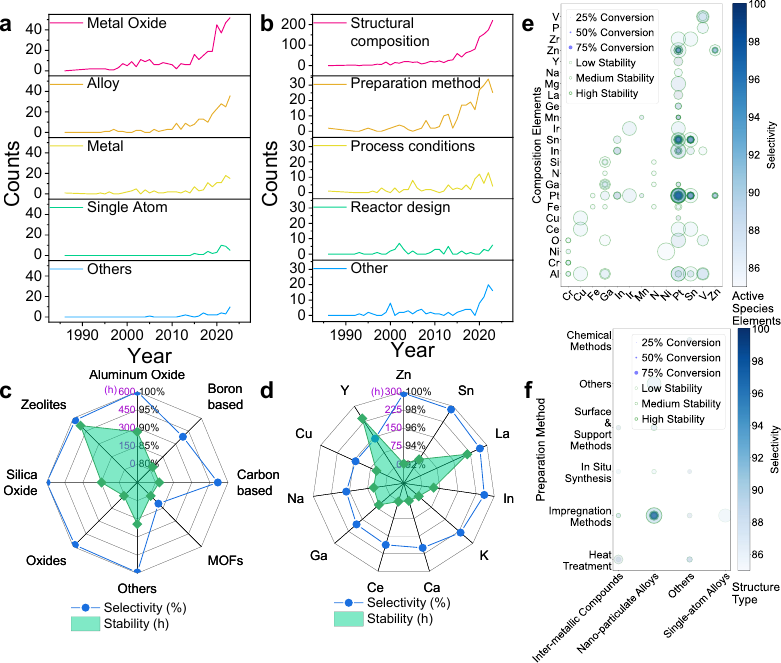}
		\caption{\textbf{Example of visual analysis results.} Line charts for annual publication numbers: \textbf{a,} different catalyst types; \textbf{b,} Performance enhancement sources. Radar charts for peak performance of single factors, with selectivity (black) and stability (purple) scales: \textbf{c,} Promoter elements; \textbf{d,} Support materials. Bubble charts for dual-variable correlations, show selectivity (color depth), conversion rate (bubble size), and stability (bubble edge thickness), aiming for high selectivity, conversion rate, and stability. Data includes only those with selectivity $\geq$85\%, conversion rate $\geq$45\%, stability $\geq$1h: \textbf{e,} Active site element-composition element; \textbf{f,} Alloy structure type-preparation method. Complete data charts are available in the SI.}
		\label{fig:3}
	\end{figure}

	For instance, publication analysis showed surging alloy research since 1995 and single-atom catalyst studies post-2015 (Fig. \ref{fig:3}a), driven by advancements in structural composition (Fig. \ref{fig:3}b). Performance analysis identified optimal promoter elements (Zn, Sn, La) (Fig. \ref{fig:3}c) and support materials (alumina, zeolites) (Fig. \ref{fig:3}d), while combination studies revealed superior performance in multi-metal systems, particularly Pt-based catalysts with Sn, Zn, In promoters (Fig. \ref{fig:3}e). Moreover, impregnation-prepared nanometallic catalysts demonstrated high conversion rates and selectivity, contrasting with single-atom alloys' high selectivity but lower conversion rates (Fig. \ref{fig:3}f).
	
	This comprehensive analysis reveals variable interactions and guides catalyst optimization, recommending Pt-based systems for selectivity and metal oxides for conversion rates, while highlighting the promise of single-atom and nanostructured catalysts. These findings not only establish performance benchmarks in PDH catalysis, but also demonstrate how our LLM-based methodology enables real-time scientific insight extraction, facilitating industrial-oriented catalyst design optimization.
		
	\subsection{Hallucination mitigation}
	
	Unlike search engines, LLMs' process of understanding information and outputting it anew provides LLMs' creativity while inevitably accompanying the "hallucination" phenomenon, referring to false information generated by LLMs without sufficient evidence support or contextually inconsistent, off-input information responses\cite{40}. LLM hallucinations mainly originate from statistical biases in the training phase, noise in training data, and decision strategies when handling uncertain or multi-interpretable information during the alignment phase\cite{40}. Currently, there is no solution within the field\cite{9,40,41}, and research suggests that hallucinations are unavoidable\cite{42,43}. Especially in specialized sub-fields, LLMs greatly exacerbate the hallucination phenomenon due to extremely scarce data exposure. For example, tests in literature\cite{44} show that even the most advanced GPT-4 only has a 73.3\% accuracy rate when answering professional multiple-choice questions, which is far from sufficient for scientific research fields that strictly require correctness. In scientific research, this may lead to unrealistic academic conclusions and misleading research directions, often meaning great waste of time and material consumption\cite{41}. Therefore, effective hallucination mitigation is crucial for ensuring the scientific nature and reliability of automated review generation.
	
	To address the challenge of hallucinations in LLMs, a high priority has been placed on the detection and prevention of such phenomena. In the entire automated review generation process, we adopted a multi-level filtering and verification quality control strategy, similar to the concept of retrieval-augmented generation (RAG)\cite{12,45}, to mitigate and correct hallucinations:
	
	\textbf{Prompt design and task decomposition.} Firstly, we utilized strict and clear text summary guiding prompts, aimed at enhancing the scientific rationality of LLM's outputs and ensuring accuracy and reliability in its analysis and generation processes. Notably, the task of automated review generation aligns well with the strengths of LLMs---information extraction and text generation capabilities. LLMs can rapidly and accurately extract core information from a vast array of literature and integrate it into a coherent and rigorous review text. To enhance efficiency and quality, we deconstructed the core of the review writing process, namely literature reading and summarization, into a series of text summarization tasks. This approach is adopted because summaries generated by LLM significantly surpass manually crafted and fine-tuned model-generated summaries in terms of fluency, factual consistency, and flexibility\cite{46}. By establishing a list of questions, we directed the model to extract relevant content from the literature and respond based on this content, subsequently conducting a comprehensive analysis of all literature citations and responses. Ultimately, the LLM generates high-quality paragraphs closely related to the topic. Additionally, we employed a single-round, segmented generation strategy to avoid truncation limitations of approximately 8K output length. By reasonably segmenting long texts for generation, we not only ensured that the output was completed in a single conversational round but also provided finer parallel granularity to improve generation efficiency. In practice, we divided the 35 questions into 5 groups, ensuring that the generation results for each group could be successfully completed within the 8K limit of the LLM. This granularity avoids efficiency drops due to a high proportion of shared content and identical prompt frameworks, thereby enhancing processing speed while ensuring the quality of text generation.
	
	\textbf{Hallucination filtering and verification.} To mitigate and rectify hallucinations, we employed a layered filtering and verification approach:
	
	\begin{enumerate}
		\item Text format filtering: Noting that hallucinations often disrupt text formatting, we applied a predefined XML format template to filter out disarrayed texts.
		
		\item DOI verification: DOIs, a combination of symbols and numbers lacking direct semantic linkage to context, present a challenge in generation and are prone to hallucinations. Yet, the precise reference nature of DOIs allows for verification. Through strict DOI verifications on generated content, we suppressed hallucinatory content from advancing further, ensuring each generated conclusion is traceable to its original source.
		
		\item Relevance verification: Within the RAG system, documents related in semantics but lacking correct answers are particularly detrimental\cite{47}. We scrutinized each response in the knowledge extraction phase to ensure its relevance, eliminating off-topic answers with relevant keywords.
		
		\item Self-consistency\cite{48} verification: For text summarization, where a definitive correct answer exists, recognizing that the stochasticity of hallucinations means correct answers should recur more frequently across iterations, we employ aggregation from repeated queries to effectively suppress hallucinations.
		
		\item Full data stream traceability mechanism: By using DOIs as key reference identifiers for each piece of generated content and mandating citations for every conclusion, we enable review readers to easily trace back to the original literature, supporting verification and deeper exploration in topics of interest.
	\end{enumerate}
	
	\textbf{Effectiveness of hallucination mitigation.} In evaluating the effectiveness of hallucination mitigation, we employed a confusion matrix to classify outcomes according to whether the LLM provided content and its pertinence to the original text, differentiating between two types of inaccuracies: false positives, which include fabricated or inconsistent information, and false negatives, referring to overlooked or partially extracted content. Our focus was primarily on reducing false positives, while adopting a relatively tolerant stance on false negatives.
	
\begin{figure}[H]
		\centering
		\includegraphics[width=1\textwidth]{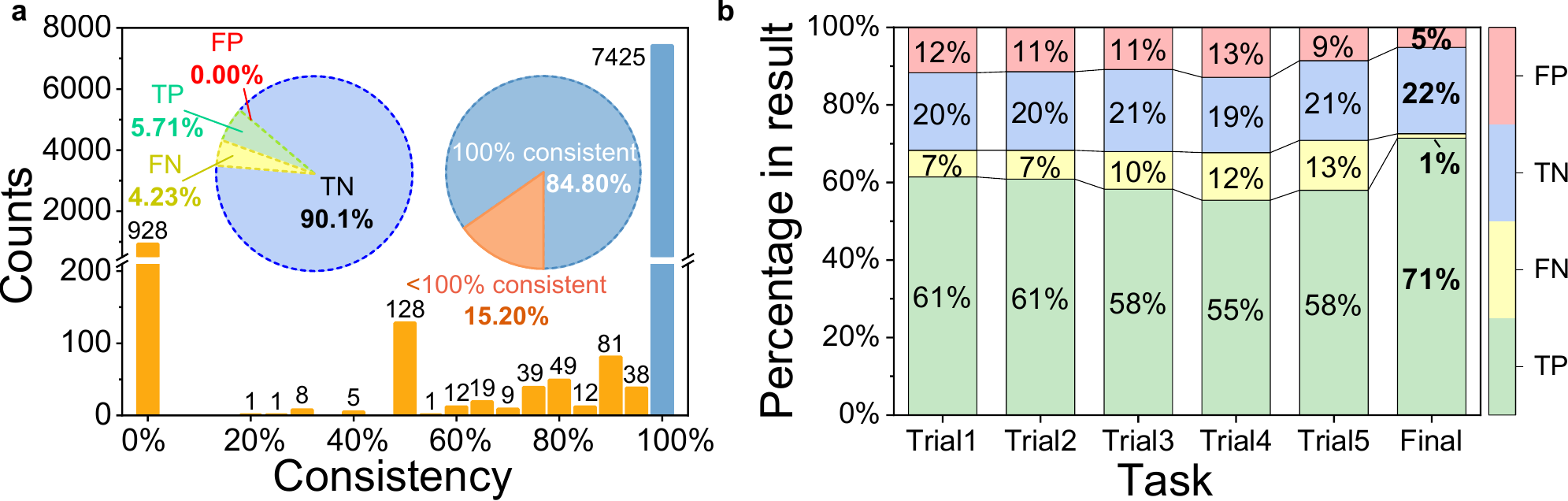}
		\caption{\textbf{Effectiveness of hallucination mitigation.} \textbf{a,} Consistency as determined by LLMs between direct LLM responses and aggregated results during the knowledge extraction phase, where blue represents 100\% consistency and orange less than 100\%. \textbf{b,} Distribution of manual sampling results for direct LLM responses and aggregated outcomes during the data mining phase, with TP (True Positive), TN (True Negative), FP (False Positive), FN (False Negative)}
		\label{fig:4}
	\end{figure}

	Substantial progress was made in mitigating hallucinations. Specifically, in the paragraph generation part, which was achieved through 9 repetitions of 35-paragraph generation tasks, a total of 315 paragraphs that passed format checking and DOI checking were needed. Throughout the entire paragraph generation process, statistics show that LLM cumulatively performed 875 generations, of which only 36\% of generation results passed after format and DOI list checks. In the analysis involving 343 topic-related literature, we divided 35 questions into 7 questions per segment for each literature, i.e., 5 segments per literature, totaling 1715 knowledge extractions. By conducting 5 repeated questions in each knowledge extraction, we obtained a total of 8575 answers, and finally aggregated 2783 effective information combinations after excluding answers unrelated to the literature and questions. Among these, up to 84.80\% of the results were judged by LLM to have a 100\% consistency rate when compared with the aggregated results (Table \ref{tab:hallucination_mitigation} and Fig. \ref{fig:4}a), thus verifying the model's stability. For specific methods, see the Methods section. This method also provides a rough standard for judging the proportion of hallucinations, which can be used in the screening and evaluation of LLMs.
		
	To assess the effectiveness of the knowledge extraction and data mining stages, we implemented a rigorous manual verification process. Specifically, 25 randomly selected articles from each stage were evaluated by a third-year PhD student specializing in PDH research. For the knowledge extraction stage, 35 segments per article were examined, totaling 875 data points. The data mining stage assessed 14 catalyst properties, including 5 direct answer repetitions and final generated results, encompassing 1750 and 350 data points respectively. We employed precise classification criteria for the evaluation. In the knowledge extraction phase, true negatives (TN) were instances where the article did not address the guiding question and the LLM correctly identified it as irrelevant. True positives (TP) occurred when relevant topics were accurately extracted. False negatives (FN) were cases where relevant topics were incompletely extracted or incorrectly deemed irrelevant. False positives (FP) included irrelevant topics mistakenly identified as relevant or extractions that exceeded or deviated from the article's actual content. Similar criteria were applied to the data mining stage, with particular attention to unit conversion errors, which were classified as false positives even if numerical values were correct. Consistency comparisons were conducted using the Claude2 model through designed prompt templates, comparing pre- and post-aggregation texts and statistically analyzing the model's scoring results. Based on these evaluations, we calculated key metrics including accuracy, false positive rate (with 95\% confidence intervals), precision, recall, F1 score, and consistency rate (Table \ref{tab:hallucination_mitigation}). Confidence intervals for false positive rates were computed using Python3's statsmodels library. For detailed results and calculation methods, see SI.
	
	It is crucial to emphasize that this manual verification step was conducted to demonstrate the method's effectiveness during the proof-of-concept phase and is not required in the actual automated review generation process. The detailed results are presented in Table \ref{tab:hallucination_mitigation}.
		
	\begin{table}[htbp]
		\centering
		\caption{Comparison of results before and after self-consistency aggregation}
		\label{tab:hallucination_mitigation}
		\small
		\footnotesize
		\setlength{\tabcolsep}{1pt}
		\begin{tabular*}{\textwidth}{@{\extracolsep{\fill}}l*{8}{>{\centering\arraybackslash}m{0.09\textwidth}}}
			\toprule
			\makecell[c]{\textbf{Stage}} & \makecell[c]{\textbf{Data}\\\textbf{Points}} & \makecell[c]{\textbf{Accu-}\\\textbf{racy}} & \makecell[c]{\textbf{False}\\\textbf{Positive}\\\textbf{Rate}} & \makecell[c]{\textbf{95\%CI}\\\textbf{of FPR}} & \makecell[c]{\textbf{Preci-}\\\textbf{sion}} & \makecell[c]{\textbf{Recall}} & \makecell[c]{\textbf{F1}\\\textbf{Score}} & \makecell[c]{\textbf{Consist}\\\textbf{-ency}} \\
			\midrule
			\multirow{3}{*}[-1ex]{\begin{tabular}[c]{@{}c@{}}Knowledge\\Extraction\\(Aggregated)\end{tabular}} & \multirow{3}{*}{875} & \multirow{3}{*}{95.77\%} & \multirow{3}{*}{0.000\%} & 0.000\% & \multirow{3}{*}{100.0\%} & \multirow{3}{*}{57.47\%} & \multirow{3}{*}{72.99\%} & \multirow{3}{*}{84.80\%} \\
			& & & & - & & & & \\
			& & & & 0.485\% & & & & \\
			\midrule
			\multirow{3}{*}[-1ex]{\begin{tabular}[c]{@{}c@{}}Data Mining\\(Direct\\Response)\end{tabular}} & \multirow{3}{*}{1750} & \multirow{3}{*}{79.09\%} & \multirow{3}{*}{35.34\%} & 31.45\% & \multirow{3}{*}{84.14\%} & \multirow{3}{*}{85.68\%} & \multirow{3}{*}{84.90\%} & \multirow{6}{*}{86.60\%} \\
			& & & & - & & & & \\
			& & & & 39.42\% & & & & \\
			\cmidrule{1-8}
			\multirow{3}{*}[-1ex]{\begin{tabular}[c]{@{}c@{}}Data Mining\\(Aggregated)\end{tabular}} & \multirow{3}{*}{350} & \multirow{3}{*}{93.71\%} & \multirow{3}{*}{18.75\%} & 12.20\% & \multirow{3}{*}{93.28\%} & \multirow{3}{*}{98.43\%} & \multirow{3}{*}{95.79\%} & \\
			& & & & - & & & & \\
			& & & & 27.70\% & & & & \\
			\bottomrule
		\end{tabular*}
	\end{table}
	
	The data comparison underscores the efficacy of self-consistency verifications, revealing a substantial decrease in hallucinations, i.e., false positive content, while also compensating for some false negatives, where information was not fully extracted (Fig. \ref{fig:4}b). In the knowledge extraction phase, critical for review content, our manual sampling found no fabricated conclusions by LLMs (Fig. \ref{fig:4}a), attesting to our method's scientific integrity and reliability. From the sampling results, we are over 95\% confident that the likelihood of hallucinations in this part is less than 0.5\% (Table \ref{tab:hallucination_mitigation}), which is also the source of our confidence that this method supports fully automated processes without manual intervention. Analysis of false positives in the post-aggregation data mining phase revealed hallucinations typically involved correct numerical extraction but with errors in units or definitions. False negatives mainly stemmed from LLMs' inability to comprehend highly abstract expressions, reflecting a general LLM's limited understanding of highly specialized scientific concepts. The incidence of hallucinations in knowledge extraction was significantly lower than in data mining, as answering questions did not involve converting units and concepts, thus avoiding the most challenging part of testing an LLM's grasp of scientific knowledge. Domain-specific models enhanced by domain-adaptive pretraining (DAPT)\cite{49} are poised to mitigate this issue. Opting not to fine-tune LLMs for specific domains in this study prioritizes out-of-the-box functionality and multi-domain generalization, utilizing a general LLM as the base. Comparisons between RAG and fine-tuning effects in specific domains indicate that RAG sustains efficacy with contextually new knowledge and offers a significantly lower initial cost\cite{50}, aligning with our objective to support researchers' entry into diverse fields efficiently.
	
	Considering the stringent accuracy requirements in research, increasing the number of repetitions can significantly reduce the probability of hallucinations appearing in aggregated results. Binomial probability calculations indicate that theoretically, a model with 79.09\% accuracy yields aggregated prediction accuracies of 93.49\%, 96.12\%, and 97.64\% after five, seven, and nine independent predictions, respectively, aligning with our sampling results (Table \ref{tab:hallucination_mitigation}). Detailed sampling outcomes and calculations are available in the SI. We believe that 5 repetitions is an ideal empirical value, and users do not need to change this parameter when using it.
	
	On this foundation, every conclusive description in the generated reviews is supported by literature references and has been verified by relevant field researchers through tracing the cited literature, confirming that all literature references are correctly linked to the original publications and that the descriptions in the generated reviews correspond to those in the original publications.
	
	This multi-layered strategy for hallucination control has built an effective verification system, ensuring the scientific integrity and reliability of the automated review generation. Furthermore, through a full data stream traceability mechanism, the authenticity and practicality of the content are further strengthened. This not only provides a secondary means of hallucination mitigation but also allows researchers to delve into original research papers for more precise and detailed academic information while accessing fast, automated research reviews. The strategy also implements a kind of literature recommendation mechanism. Since each content segment includes related DOIs, researchers can quickly locate specific original literature based on their interests and research needs, enabling deeper academic exploration.
	
	While both our method and RAG utilize LLM's context learning ability, our approach fundamentally differs by achieving systematic knowledge reconstruction through multi-stage processing rather than simple retrieval combination. This method simulates the complete academic research process and produces coherent knowledge frameworks aligned with scholarly thinking, surpassing traditional RAG's question-answering paradigm through comprehensive quality control and hallucination mitigation mechanisms.
	
	\section{CONCLUSIONS}
	
	In this study, we introduce an innovative LLM-based automated review generation method, addressing two fundamental challenges in scientific research: improving literature review efficiency and mitigating LLM hallucination risks. Through proposing an evaluation framework that ensures the objectivity and reliability via statistical validation, and innovatively compared LLM-generated reviews with high-quality manual reviews, we demonstrate that our modular end-to-end approach produces reviews comparable to or exceeding human-written ones, while maintaining high reliability and traceability. Expert evaluation using PDH catalysts as a case study confirms the method's effectiveness: generated reviews are comparable to manual reviews in length and citations, show no hallucinations, and have impeccable reference accuracy. Statistical validation confirms the method's effectiveness in hallucination reduction, with testing on 875 LLM outputs from 25 random articles showing hallucination probability below 0.5\% at 95\% confidence. The quality assurance pipeline ensures robust data processing. Additionally, our advanced data mining module offers experienced users' in-depth field integrated perception, fully exploiting LLMs' analytical capabilities. Furthermore, an open-source user-friendly one-click program developed for Windows platforms significantly simplifies the review generation process.
	
	The method's architecture offers significant advantages through its cross-disciplinary applicability without manual intervention or domain-specific knowledge injection. Its modular design enables component reuse for literature tracking, topic discovery, and dataset construction, while achieving cross-disciplinary generalization through LLM's inherent contextual adaptability. This means that by providing corresponding domain literature input, the method can generate high-quality reviews across various disciplines. Future development will focus on enhanced multimodal processing capabilities, automated scientific question generation and answering, personalized text generation, integration with existing academic tools, and domain-specific features for structured data analysis.
	
	This advancement heralds a new era in human-machine academic collaboration, offering broad prospects for LLMs as writing assistance tools. While not intended to replace traditional manual reviews, our method serves as a powerful auxiliary tool for rapid domain overview and research hotspot identification, laying the foundation for in-depth analysis. Beyond its demonstrated excellence in chemistry, the method's technical framework exhibits remarkable cross-disciplinary applicability, potentially breaking down barriers between fields and catalyzing interdisciplinary innovation. By revolutionizing researcher-literature interaction and accelerating knowledge dissemination, this milestone advancement holds profound implications for knowledge base construction, literature recommendation, and structured academic writing, heralding a new era of scientific research productivity and interdisciplinary collaboration.
	
	\section{METHODS}
	
\begin{figure}[H]
		\centering
		\includegraphics[width=1\textwidth]{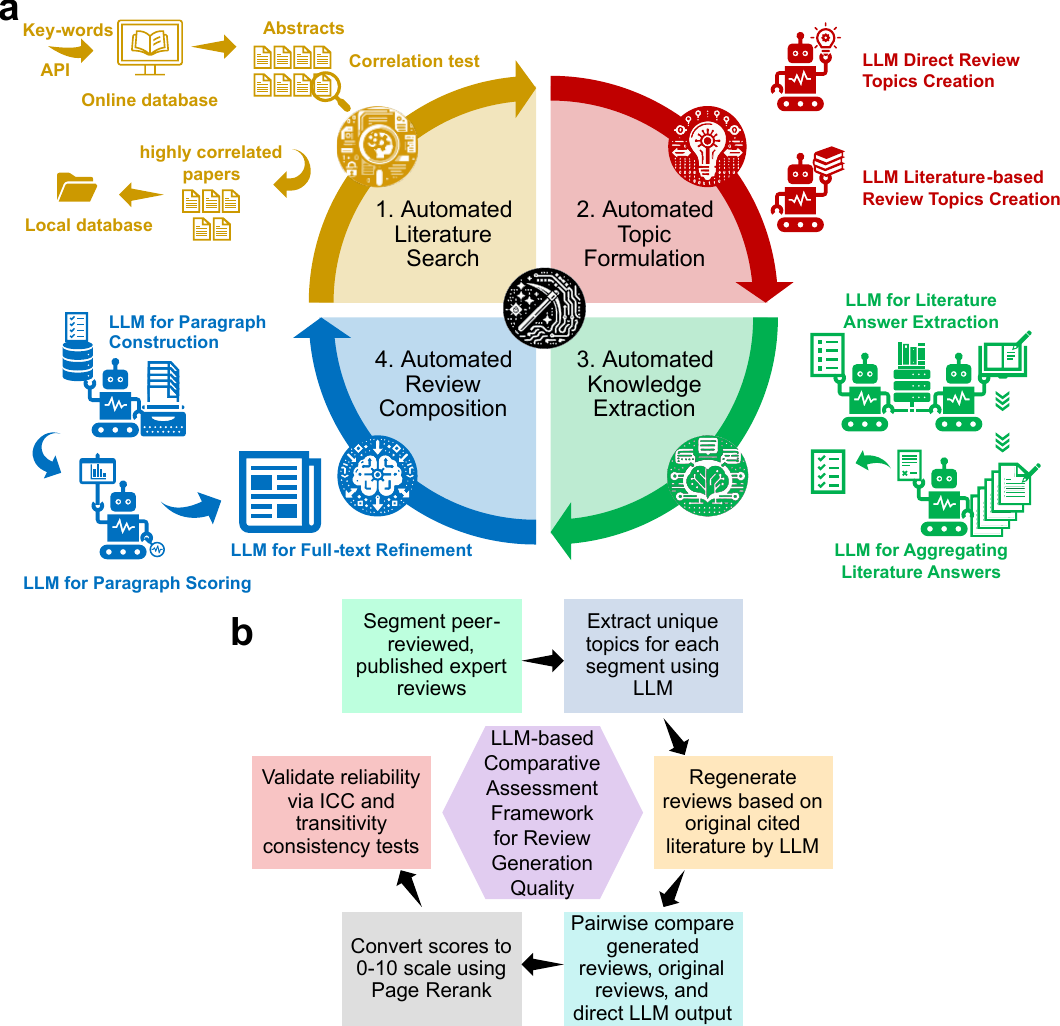}
		\caption{\textbf{a, Flowchart of the automated review generation method based on large language models.} It includes four modules: i) literature search, ii) topic formulation, iii) knowledge extraction, iv) review composition, as well as an additional data mining module. \textbf{b, Flowchart of the quality assessment framework for review generation based on large language models.}}
		\label{fig:5}
	\end{figure}

	Our method consists of four core components: literature search, topic formulation, knowledge extraction and review composition, along with a data mining module (Fig. \ref{fig:5}a) and quality assessment framework (Fig. \ref{fig:5}b). All prompt templates are available in SI and GitHub without requiring user adjustment.
		
	\subsection{Literature search}
	
	Literature retrieval begins with journal selection from journal classification tables, followed by an API-based keyword search and preliminary title/abstract filtering using a keyword list, with review-type literature marked separately (Fig. \ref{fig:5}a, i).
	
	\subsection{Topic formulation}
	
	Review topics can be constructed either through direct LLM outline drafting or through LLM extraction and refinement of existing literature reviews (Fig. \ref{fig:5}a, ii). After obtaining a list of topics, additional topics can be manually added and sorted as needed. This manual addition is not mandatory but provides an interface for advanced users to intervene if necessary.
	
	\subsection{Knowledge extraction}
	
	Based on the topic list, LLM generates extraction questions and conducts multiple rounds of information retrieval from each article. The LLM evaluates answer relevance to questions using structured prompts, where combinations of questions, LLM-aggregated relevant answers, and their corresponding citations constitute valid information combinations for subsequent processing. For literature exceeding the context window of LLM, the text is segmented into approximately equal parts, processed separately, and results are integrated during answer aggregation (Fig. \ref{fig:5}a, iii).
	
	\subsection{Review composition}
	
	Extracted answers are associated with source DOIs and integrated into topic-specific paragraphs. Through multiple iterations and LLM scoring, optimal versions are selected to form the preliminary draft, followed by citation verification and format standardization. For answers exceeding context window length, LLM performs compression based on referenced texts and extracted answers until fitting within window limits (Fig. \ref{fig:5}a, iv).
	
	\subsection{Data mining}
	
	The data mining module extends knowledge extraction (Fig. \ref{fig:5}a, iii) capabilities for specific data extraction and aggregation, enables extraction of user-defined targets (e.g., catalyst types, compositions, performance metrics) from literature. The LLM performs multiple rounds of parsing and extraction in XML format, followed by result aggregation. The extracted data undergoes standardization and cleaning, with GPT4-generated code facilitating statistical analysis and visualization, requiring no programming expertise from users.
	
	\subsection{Quality Assessment}
	
	The evaluation framework employs dual baselines using manual Q1 journal reviews and direct LLM generation for quality assessment (Fig. \ref{fig:5}b). High-quality reviews are semantically segmented, with corresponding content regenerated using our method and compared against direct LLM-generated content. Assessment utilizes chain-of-thought prompts across 27 scoring items in 9 categories, implementing cross-evaluation and repetition strategies to mitigate bias. The page rerank algorithm converts relative comparisons to absolute scores on a 0-10 scale, with framework reliability validated through intraclass correlation coefficient (ICC) tests and transitive consistency ratio (TCR) analyses.
	
	\section{ACKNOWLEDGMENT}
	
	We acknowledge the Natural Science Foundation of China (No. 22121004), the Haihe Laboratory of Sustainable Chemical Transformations, the Program of Introducing Talents of Discipline to Universities (BP0618007) and the XPLORER PRIZE for financial support. We also acknowledge generous computing resources at High Performance Computing Center of Tianjin University.
	
	\section{AUTHOR CONTRIBUTIONS}
	
	J.G and Z.Z. conceived and supervised the project. S.W. and X.M. designed the research and developed the program. S.W. led the manuscript preparation. D.L. performed manual verification of hallucination. L.L., X.S., and X.C. advanced the integration of computational models, while X.L., R.L., C.P. and C.D. advanced the catalytic science. All authors contributed to writing and revising the manuscript.
	
	\section{DATA AVAILABILITY}
	
	Our study leverages a dataset compiled from scientific literature acquired through our institution's subscription. Due to copyright considerations, the dataset itself cannot be made publicly available. However, we ensure that our research's integrity and reproducibility do not rely on direct access to these proprietary documents. Instead, we provide extensive documentation on the dataset's structure, the criteria used for literature selection, and the analysis methods applied, enabling interested researchers to reconstruct a similar dataset from publicly available resources or their institutional subscriptions.
	
	Furthermore, to facilitate a deeper understanding of our research process and promote further exploration and innovation, we have made all intermediate data, excluding the copyrighted full-text articles, publicly available on GitHub [https://github.com/TJU-ECAT-AI/AutomaticReviewGenerationData]. This repository includes the prompts used in our study and the corresponding responses generated by the large language model. By sharing these resources, we aim to provide valuable insights into our methodology and encourage other researchers to build upon our work, advancing the field of natural language processing and its applications in scientific literature analysis.
	
	\section{CODE AVAILABILITY}
	
	The custom code developed for this research is central to our conclusions and is made available to ensure transparency and reproducibility of our results. The codebase, including all relevant custom scripts and mathematical algorithms, has been open-sourced under the Apache 2.0 license and is accessible via our GitHub repository at [https://github.com/TJU-ECAT-AI/AutomaticReviewGeneration]. We encourage users to review the license for any usage restrictions that may apply.
	
	As stated in the text, all LLMs invoked in this article are Claude 2, except for the Evaluation of generated review quality section which uses Claude 3.5 Sonnet, Qwen2-72b-Instruct, and Qwen2-7b-Instruct. It is important to note that our data processing work, except for the evaluation section, was completed prior to November 21, 2023, utilizing the then-latest available Claude 2 API version [https://www.anthropic.com/news/claude-2, https://www.anthropic.com/news/claude-2-1]. Anthropic did not publish specific minor version numbers within the Claude 2 series, only distinguishing between Claude 2, Claude 2.1, and the subsequent Claude 3 series. Our proposed framework demonstrates good adaptability, with overall effectiveness increasing as the performance of the underlying model improves. This characteristic has been amply demonstrated in our evaluation work, indicating that the framework's efficacy is not strictly dependent on any particular model version. The use of different LLMs in the Evaluation of generated review quality section was primarily to assess the performance of the latest and most powerful open-source and closed-source models (as of September 2024) under the method described in this paper.
	
	Our published graphical user interface (GUI) leverages certain APIs for functionality, which, due to legal and regulatory requirements, necessitate that users provide their own API keys. This requirement is detailed in the documentation accompanying the code repository to assist users in setting up and utilizing the GUI effectively.
	
	\section{CONFLICT OF INTEREST STATEMENT}
	
	The authors declare no competing interests.
	
	\section{ADDITIONAL INFORMATION}
	
	Supplementary information is available for this paper.
	
	Correspondence and requests for materials should be addressed to J.G.


\begin{thebibliography}{50}
		\bibitem{1} Ermel APC, Lacerda DP, Morandi MIW, Gauss L. \textit{Literature reviews: modern methods for investigating scientific and technological knowledge}. Springer Nature; 2021.
		
		\bibitem{2} Lawrence S. Free online availability substantially increases a paper's impact. \textit{Nature}. May 31 2001;411(6837):521.
		
		\bibitem{3} Lok C. Speed reading: scientists are struggling to make sense of the expanding scientific literature. Corie Lok asks whether computational tools can do the hard work for them. \textit{Nature}. 2010;463(7280):416-419.
		
		\bibitem{4} Wu PF, Zhang F. Recent Advances in Lead Chemisorption for Perovskite Solar Cells. \textit{Transactions of Tianjin University}. Oct 2022;28(5):341-357.
		
		\bibitem{5} Khurana D, Koli A, Khatter K, Singh S. Natural language processing: state of the art, current trends and challenges. \textit{Multimed Tools Appl}. 2023;82(3):3713-3744.
		
		\bibitem{6} Liu Y, Han T, Ma S, et al. Summary of chatgpt-related research and perspective towards the future of large language models. \textit{Meta-Radiology}. 2023:100017.
		
		\bibitem{7} Hendrycks D, Burns C, Basart S, et al. Measuring massive multitask language understanding. \textit{arXiv preprint arXiv:200903300}. 2020;
		
		\bibitem{8} Rein D, Hou BL, Stickland AC, et al. Gpqa: A graduate-level google-proof q\&a benchmark. \textit{arXiv preprint arXiv:231112022}. 2023;
		
		\bibitem{9} White AD. The future of chemistry is language. \textit{Nature Reviews Chemistry}. Jul 2023;7(7):457-458.
		
		\bibitem{10} Lála J, O'Donoghue O, Shtedritski A, Cox S, Rodriques SG, White AD. Paperqa: Retrieval-augmented generative agent for scientific research. \textit{arXiv preprint arXiv:231207559}. 2023;
		
		\bibitem{11} Skarlinski MD, Cox S, Laurent JM, et al. Language agents achieve superhuman synthesis of scientific knowledge. \textit{arXiv preprint arXiv:240913740}. 2024;
		
		\bibitem{12} Lewis P, Perez E, Piktus A, et al. Retrieval-augmented generation for knowledge-intensive nlp tasks. \textit{Advances in Neural Information Processing Systems}. 2020;33:9459-9474.
		
		\bibitem{13} Wei S, Xu X, Qi X, et al. AcademicGPT: Empowering Academic Research. \textit{arXiv preprint arXiv:231112315}. 2023;
		
		\bibitem{14} Yang Z, Zhu Z. CuriousLLM: Elevating Multi-Document QA with Reasoning-Infused Knowledge Graph Prompting. \textit{arXiv preprint arXiv:240409077}. 2024;
		
		\bibitem{15} Ma C, Zhang WE, Guo M, Wang H, Sheng QZ. Multi-document summarization via deep learning techniques: A survey. \textit{Acm Comput Surv}. 2022;55(5):1-37.
		
		\bibitem{16} Nikiforovskaya A, Kapralov N, Vlasova A, Shpynov O, Shpilman A. Automatic generation of reviews of scientific papers. IEEE; 2020:314-319.
		
		\bibitem{17} Mohammad S, Dorr B, Egan M, et al. Using citations to generate surveys of scientific paradigms. 2009:584-592.
		
		\bibitem{18} Agarwal N, Reddy RS, Kiran G, Rose C. Towards multi-document summarization of scientific articles: making interesting comparisons with SciSumm. 2011:8-15.
		
		\bibitem{19} Jaidka K, Khoo C, Na J-C. Deconstructing human literature reviews--a framework for multi-document summarization. 2013:125-135.
		
		\bibitem{20} Kasanishi T, Isonuma M, Mori J, Sakata I. SciReviewGen: A Large-scale Dataset for Automatic Literature Review Generation. \textit{arXiv preprint arXiv:230515186}. 2023;
		
		\bibitem{21} Sami AM, Rasheed Z, Kemell K-K, et al. System for systematic literature review using multiple ai agents: Concept and an empirical evaluation. \textit{arXiv preprint arXiv:240308399}. 2024;
		
		\bibitem{22} Agarwal S, Laradji IH, Charlin L, Pal C. LitLLM: A Toolkit for Scientific Literature Review. \textit{arXiv preprint arXiv:240201788}. 2024;
		
		\bibitem{23} Haryanto CY. LLAssist: Simple Tools for Automating Literature Review Using Large Language Models. \textit{arXiv preprint arXiv:240713993}. 2024;
		
		\bibitem{24} Joos L, Keim DA, Fischer MT. Cutting Through the Clutter: The Potential of LLMs for Efficient Filtration in Systematic Literature Reviews. \textit{arXiv preprint arXiv:240710652}. 2024;
		
		\bibitem{25} Li Y, Chen L, Liu A, Yu K, Wen L. ChatCite: LLM Agent with Human Workflow Guidance for Comparative Literature Summary. \textit{arXiv preprint arXiv:240302574}. 2024;
		
		\bibitem{26} Gilardi F, Alizadeh M, Kubli M. ChatGPT outperforms crowd workers for text-annotation tasks. \textit{P Natl Acad Sci USA}. Jul 25 2023;120(30):e2305016120.
		
		\bibitem{27} Törnberg P. Chatgpt-4 outperforms experts and crowd workers in annotating political twitter messages with zero-shot learning. \textit{arXiv preprint arXiv:230406588}. 2023;
		
		\bibitem{28} Zhang X, Li Y, Wang J, et al. Large Language Models as Evaluators for Recommendation Explanations. 2024:33-42.
		
		\bibitem{29} Kadavath S, Conerly T, Askell A, et al. Language models (mostly) know what they know. \textit{arXiv preprint arXiv:220705221}. 2022;
		
		\bibitem{30} Zheng LM, Chiang WL, Sheng Y, et al. Judging LLM-as-a-Judge with MT-Bench and Chatbot Arena. \textit{Adv Neur In}. 2023;36:46595-46623.
		
		\bibitem{31} Bai Y, Ying J, Cao Y, et al. Benchmarking foundation models with language-model-as-an-examiner. \textit{Advances in Neural Information Processing Systems}. 2024;36
		
		\bibitem{32} Gou Z, Shao Z, Gong Y, et al. Critic: Large language models can self-correct with tool-interactive critiquing. \textit{arXiv preprint arXiv:230511738}. 2023;
		
		\bibitem{33} Gao M, Hu X, Ruan J, Pu X, Wan X. Llm-based nlg evaluation: Current status and challenges. \textit{arXiv preprint arXiv:240201383}. 2024;
		
		\bibitem{34} Wang P, Li L, Chen L, et al. Large language models are not fair evaluators. \textit{arXiv preprint arXiv:230517926}. 2023;
		
		\bibitem{35} Liu Y, Iter D, Xu Y, Wang S, Xu R, Zhu C. G-eval: Nlg evaluation using gpt-4 with better human alignment. \textit{arXiv preprint arXiv:230316634}. 2023;
		
		\bibitem{36} Li Z, Wang C, Ma P, et al. Split and merge: Aligning position biases in large language model based evaluators. \textit{arXiv preprint arXiv:231001432}. 2023;
		
		\bibitem{37} Shen C, Cheng L, Nguyen X-P, You Y, Bing L. Large language models are not yet human-level evaluators for abstractive summarization. \textit{arXiv preprint arXiv:230513091}. 2023;
		
		\bibitem{38} Gong J, Hou SX, Wang Y, Ma XB. Progress in Processes and Catalysts for Dehydrogenation of Cyclohexanol to Cyclohexanone. \textit{Transactions of Tianjin University}. Jun 2023;29(3):196-208.
		
		\bibitem{39} Lin XY, Zhen SY, Wang XH, et al. Data-Driven Design of Single-Atom Electrocatalysts with Intrinsic Descriptors for Carbon Dioxide Reduction Reaction. \textit{Transactions of Tianjin University}. Oct 2024;30(5):459-469.
		
		\bibitem{40} Zhang Y, Li Y, Cui L, et al. Siren's Song in the AI Ocean: A Survey on Hallucination in Large Language Models. \textit{arXiv preprint arXiv:230901219}. 2023;
		
		\bibitem{41} Sanderson K. GPT-4 is here: what scientists think. \textit{Nature}. Mar 30 2023;615(7954):773-773.
		
		\bibitem{42} Xu Z, Jain S, Kankanhalli M. Hallucination is inevitable: An innate limitation of large language models. \textit{arXiv preprint arXiv:240111817}. 2024;
		
		\bibitem{43} Kalai AT, Vempala SS. Calibrated language models must hallucinate. 2024:160-171.
		
		\bibitem{44} Wu S, Koo M, Blum L, et al. A comparative study of open-source large language models, gpt-4 and claude 2: Multiple-choice test taking in nephrology. \textit{arXiv preprint arXiv:230804709}. 2023;
		
		\bibitem{45} Truhn D, Reis-Filho JS, Kather JN. Large language models should be used as scientific reasoning engines, not knowledge databases. \textit{Nature Medicine}. 2023:1-2.
		
		\bibitem{46} Pu X, Gao M, Wan X. Summarization is (almost) dead. \textit{arXiv preprint arXiv:230909558}. 2023;
		
		\bibitem{47} Cuconasu F, Trappolini G, Siciliano F, et al. The power of noise: Redefining retrieval for rag systems. 2024:719-729.
		
		\bibitem{48} Wang X, Wei J, Schuurmans D, et al. Self-consistency improves chain of thought reasoning in language models. \textit{arXiv preprint arXiv:220311171}. 2022;
		
		\bibitem{49} Gururangan S, Marasović A, Swayamdipta S, et al. Don't stop pretraining: Adapt language models to domains and tasks. \textit{arXiv preprint arXiv:200410964}. 2020;
		
		\bibitem{50} Gupta A, Shirgaonkar A, Balaguer AdL, et al. RAG vs Fine-tuning: Pipelines, Tradeoffs, and a Case Study on Agriculture. \textit{arXiv preprint arXiv:240108406}. 2024;
		
	\end{thebibliography}
	\end{document}